\documentclass[conference]{IEEEtran}
\IEEEoverridecommandlockouts
\usepackage{cite}
\usepackage{amsmath,amssymb,amsfonts}
\usepackage{algorithmic}
\usepackage{graphicx}
\usepackage{textcomp}
\usepackage{booktabs}
\usepackage{multirow}
\usepackage{xcolor}
\usepackage{hhline}
\usepackage{boldline}
\usepackage{url}

\def\BibTeX{{\rm B\kern-.05em{\sc i\kern-.025em b}\kern-.08em
    T\kern-.1667em\lower.7ex\hbox{E}\kern-.125emX}}

\title{KPCA-CAM: Visual Explainability of Deep Computer Vision Models using Kernel PCA}

\author{\IEEEauthorblockN{Sachin Karmani, Thanushon Sivakaran, Gaurav Prasad, Mehmet Ali, Wenbo Yang, Sheyang Tang}
\IEEEauthorblockA{\textit{Department of Electrical and Computer Engineering} \\
\textit{University of Waterloo}\\
Waterloo, Canada \\
\{skarmani, tsivakar, g3prasad, mapelit, paul.yang, sheyang.tang\}@uwaterloo.ca}
}

\begin{document}

\maketitle

\begin{abstract}
Deep learning models often function as black boxes, providing no straightforward reasoning for their predictions. This is particularly true for computer vision models, which process tensors of pixel values to generate outcomes in tasks such as image classification and object detection. To elucidate the reasoning of these models, class activation maps (CAMs) are used to highlight salient regions that influence a model's output. This research introduces KPCA-CAM, a technique designed to enhance the interpretability of Convolutional Neural Networks (CNNs) through improved class activation maps. KPCA-CAM leverages Principal Component Analysis (PCA) with the kernel trick to capture nonlinear relationships within CNN activations more effectively. By mapping data into higher-dimensional spaces with kernel functions and extracting principal components from this transformed hyperplane, KPCA-CAM provides more accurate representations of the underlying data manifold. This enables a deeper understanding of the features influencing CNN decisions. Empirical evaluations on the ILSVRC dataset across different CNN models demonstrate that KPCA-CAM produces more precise activation maps, providing clearer insights into the model's reasoning compared to existing CAM algorithms. This research advances CAM techniques, equipping researchers and practitioners with a powerful tool to gain deeper insights into CNN decision-making processes and overall behaviors.
\end{abstract}


\section{Introduction}

In recent years, Convolutional Neural Networks (CNNs) \cite{cnns} have emerged as the cornerstone of various artificial intelligence applications, demonstrating unparalleled performance in computer vision tasks. However, despite their remarkable achievements, CNNs are often characterized as opaque "black boxes," leaving developers in the dark regarding the underlying decision-making processes. This lack of interpretability poses significant challenges, particularly in fields where accountability, fairness, and trust are paramount.

The opacity of CNNs \cite{cnns} stems from their complex architectures, consisting of multiple layers of interconnected neurons that transform input data into high-level representations. While these networks excel at capturing intricate patterns and features, understanding how they arrive at their predictions remains elusive. This inherent black-box nature hinders not only users' ability to interpret and trust CNN outputs but also limits the potential for identifying biases, errors, or unethical decision-making within these models.

\begin{figure}[t]
\centerline{\includegraphics[height=5cm]{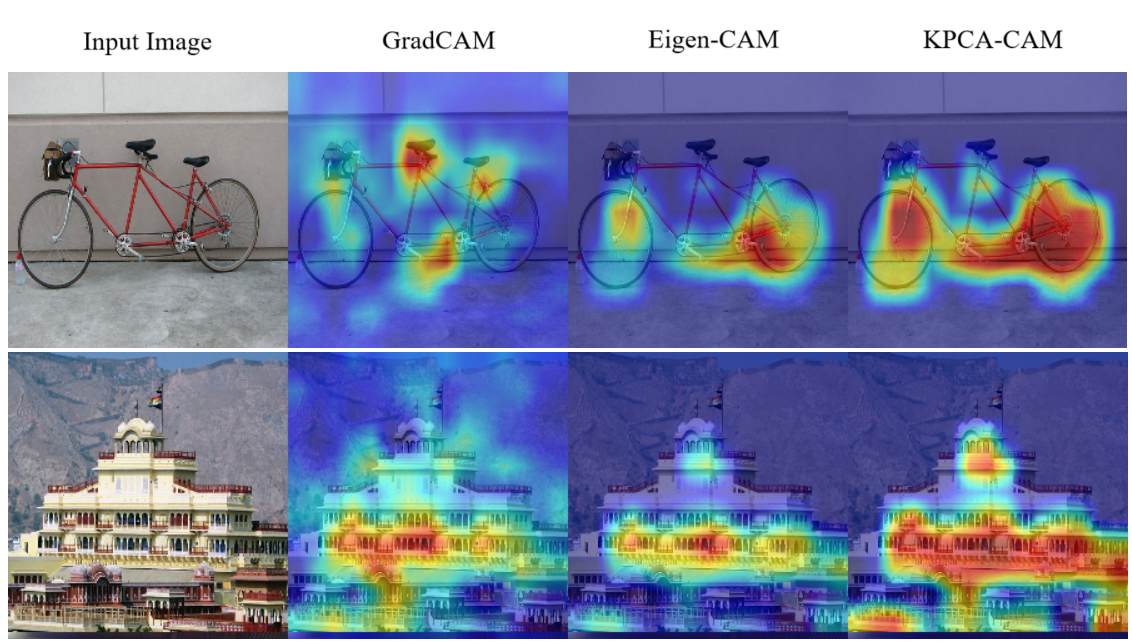}}
\caption{Comparison of existing CAM techniques with our proposed KPCA-CAM algorithm }
\label{figure_example_cams}
\end{figure}

Researchers have been addressing the challenge of interpretability in CNNs by developing techniques to visualize the input features which influenced the output of the model. Early methods like Class Activation Mapping (CAM) \cite{cam} provided insights but lacked precise localization. Gradient-weighted Class Activation Mapping (Grad-CAM) \cite{grad_cam} improved upon CAM by incorporating gradient information for finer localization. Despite further refinements like Grad-CAM++ \cite{grad_cam_++}, Score-CAM \cite{score_cam}, and Layer-CAM \cite{layer_cam}, challenges remained, prompting the development of Eigen-CAM \cite{eigen_cam}. Eigen-CAM utilizes PCA \cite{pca} to analyze CNN weight matrices, revealing critical features in the input data without requiring architectural changes \cite{cam} or gradients \cite{grad_cam} \cite{grad_cam_++}.

In this paper, we improve the CAMs produced by Eigen-CAM by proposing KPCA-CAM as seen in Fig. \ref{figure_example_cams}. This novel approach uses Kernel PCA \cite{kernelpca} instead of the standard PCA technique used in Eigen-CAM. Kernel PCA \cite{kernelpca} extends traditional PCA by employing kernel functions to map data into higher-dimensional spaces, where nonlinear relationships can be better captured. By applying Kernel PCA to the feature maps, we can effectively capture the nonlinear and complex structures inherent in CNN activations. This allows us to derive more accurate representations of the underlying data manifold, enabling a better understanding of the features influencing CNN decisions.

The major contributions of the proposed KPCA-CAM are as follows:

\begin{itemize}
    \item We showcase an intuitive CAM generation algorithm that captures the non-linearity and overall complex structures of activations without relying on gradients to produce improved activation maps over its predecessors.

    \item We demonstrate that different kernel functions used in the proposed KPCA-CAM can capture different aspects of the underlying data manifold, leading to a more diverse and comprehensive representation of the complex relationships within CNN feature maps.
\end{itemize}

\section{Background}
\label{sec:related_works}

The original CAM technique \cite{cam} represents a significant leap in the realm of CNN \cite{cnns} interpretability. These techniques offer a window into understanding which regions of an input image play pivotal roles in the decision-making process of CNNs. The foundational work by Zhou et al. introduced CAM \cite{cam}, a technique that directly visualizes feature activations related to specific output classes, but required extensive modifications to the network architecture. This method facilitated comprehension of the model's focus areas, laying the groundwork for subsequent advancements.

Building upon CAM, Selvaraju et al. introduced Grad-CAM \cite{grad_cam}, which utilized gradient information from the last convolutional layer to generate more effective visual explanations. Unlike CAM \cite{cam}, Grad-CAM's versatility allowed it to be applied across various CNN models \cite{cnns}, without changing the architecture of the model. Grad-CAM's limitations in localization granularity and interpretability for multiple objects led to refinements proposed in Grad-CAM++ \cite{grad_cam_++}, which improved the visualization by considering the importance of each pixel for the decision of interest. 

Score-CAM \cite{score_cam} utilized a score-based weighting approach to generate class activation maps. This method did not rely on gradient information, offering a more intuitive visualization of model decision-making processes. There are set of challenges, such as increased computational intensity due to need for multiple forward passes through the network and potential ambiguity in cases of subtle feature influences. Jiang et al. proposed Layer-CAM \cite{layer_cam}, enhancing the granularity of class activation mappings by leveraging hierarchical information across different CNN layers. This technique provided more detailed visual explanations of model behavior by dynamically weighting the contribution of each layer's feature maps to the final decision. Muhammed et al. introduced Eigen-CAM \cite{eigen_cam}, which employed PCA \cite{pca} on feature maps to generate class activation maps. This model-agnostic technique highlighted discriminative features with minimal computational overhead. Despite its advantages, Eigen-CAM's simplification via PCA might overlook critical details in understanding the model's decision-making process.

\section{Proposed Approach}
\label{sec:proposed_approach}

Images have instrinsic dimensionality \cite{intrinsic_dim}, composed of signals with varying degrees of freedom and cannot be evaluated using linear operators. Therefore, CNNs employ non-linear activation functions \cite{nonlinearAF} to learn abstract features in non-linear space. A limitation of Eigen-CAM is that it computes a linear combination of activations from a specific convolutional layer to provide salient features in the direction of the first principal component. While this approach offers valuable insights into the most influential features within the input data, it may oversimplify the decision-making process of the CNN due to capturing only linear combinations. CNNs operate through nonlinear transformations, meaning that the relationship between input features and network activations is often complex and nonlinear. It's reliance on linear combinations may fail to capture the full complexity of these relationships, potentially leading to a loss of important information such as the non-linear spatial relationships present within CNN feature maps. CNNs learn hierarchical representations of the input data, where features at higher layers represent increasingly abstract concepts. Eigen-CAM's linear approach may overlook these hierarchical structures, limiting its ability to provide a comprehensive understanding of CNN behavior.

\subsection{Kernel PCA}
\label{sec:kernel-pca}

PCA \cite{pca} is highly effective for linear data transformations but it may not capture the data in the underlying structure of nonlinear datasets. Kernel Principal Component Analysis (Kernel PCA) \cite{kernelpca} is an extension of PCA \cite{pca} that allows for the nonlinear mapping of data into higher-dimensional spaces, where nonlinear relationships can be better captured. Unlike PCA, Kernel PCA employs kernel functions to project data into higher-dimensional feature spaces, enabling the extraction of nonlinear patterns and structures from the data. In short, Kernel PCA \cite{kernelpca} takes a dataset and maps it into some higher dimension, and then performs PCA on the new dimensional space.

Let \( \mathbf{K} \) denote the kernel matrix, where each element represents the dot product of one point with respect to all other points. The kernel formulation of PCA computes the projections of data onto the principal components (not the components themselves). The projection from a point in the feature space onto the \( k \)-th principal component is defined as

\begin{equation}
\text{Projection}(\mathbf{x}, k) = \mathbf{v}^k \cdot \mathbf{K}.
\label{projection}
\end{equation}

In \eqref{projection}, \( \mathbf{v}^k \) is the \( k \)-th eigenvector of the kernel matrix, and \( \mathbf{K} \) is the dot product of \( \mathbf{x} \) with all other transformed points.

\subsection{Kernels}
\label{sec:kernels}

\subsubsection{Radial Basis Function (RBF)}
\label{sec:rbf}

The Radial Basis Function (RBF) \cite{kernelpca} is a kernel function that computes the similarity or distance between pairs of data points based on their radial distance from a center. The RBF kernel between two data points \( \mathbf{x}_i \) and \( \mathbf{x}_j \) is defined as

\begin{equation}
 \mathbf{K}(\mathbf{x}_i, \mathbf{x}_j) = \exp\left(-\gamma*\|\mathbf{x}_i - \mathbf{x}_j\|^2\right).\label{rbf}
\end{equation}

In \eqref{rbf}, \( \gamma \) defines the spread of the kernel and determines the influence of nearby points on the similarity measure. 

\subsubsection{Sigmoid Kernel}
\label{sec:sigmoid}

The Sigmoid function \cite{kernelpca} is an activation function commonly used to add non-linearity into a model's architecture. It converts the input to a value between 0 and 1, making it usable for binary classification tasks. It is defined between two data points \( \mathbf{x}_i \) and \( \mathbf{x}_j \) as

\begin{equation}
\mathbf{K}(\mathbf{x}_i, \mathbf{x}_j) = \tanh(\gamma*\mathbf{x}_i\mathbf{x}_j + r).
\label{sigmoid}
\end{equation}

In \eqref{sigmoid}, \( \gamma \) is a scale parameter that controls the slope of the hyperbolic tangent function, while \( r\) is an offset parameter that shifts the input to the hyperbolic tangent function.

\subsection{KPCA-CAM}
\label{sec:kpcacam}

To generate activation maps which capture non-linear representations learned by the CNN model, we apply Kernel PCA on the feature maps generated by convolutional layers. 

Let \({C}\) represent the output of the last convolution layer. We construct a kernel matrix \(\mathbf{K}\) using a kernel function applied to the feature vectors in \({C}\) to calculate pairwise distances.

The eigenvectors of \(\mathbf{K}\) can be obtained by solving:

\begin{equation}
\mathbf{K} = V\Lambda\\V^{-1}, \label{eigenvectors}
\end{equation}
where \(V\) is an orthogonal matrix and \(\Lambda\) is he diagonal matrix of eigenvalues.

The class activation map, \({L}\) is given by the projection
of  \(\mathbf{K}\) on the first eigenvector \({V_1}\) in the matrix V. Formally, this is defined as

\begin{equation}
L = \mathbf{K}V_1. \label{cam}
\end{equation}

\begin{figure}[b]
\centerline{\includegraphics[width=7cm,height=4cm]{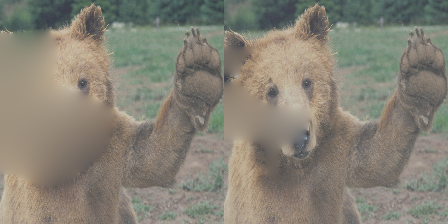}}
\caption{Using the MoRF metric \cite{road} to evaluate the confidence drop when removing 25\% (left) vs 10\% (right) of the most relevant pixels as decided by the CAM produced \cite{jacobgilpytorchcam}}
\label{figure_road_bear}
\end{figure}

\section{Experiment Results}
\label{sec:experiment_results}

In this section, we will briefly explain the experiments and metrics used to evaluate the performance of the novel approach as well as provide the performance comparison and analysis to alternate CAM methods.
We create two KPCA-CAMs, KPCA-CAM(S), which uses the sigmoid kernel and KPCA-CAM(R), which uses the RBF kernel. For the sigmoid kernel, we use a gamma value of 0.1, while we employ a gamma value of 0.001 for the RBF kernel.
Two experiments were conducted on 5000 randomly sampled datapoints from the ILSVRC validation set \cite{ilsvrc}, which contains images along with their classes and bounding boxes. We used VGG-16 \cite{vgg16}, ResNet-50  \cite{resnet50} and DenseNet-161  \cite{desnenet161} as the base models and generated class activation maps from the last convolutional layer of these models.

\subsubsection{Weakly supervised object localization}
\label{sec:localization}

\begin{figure*}[t]
\centerline{\includegraphics[width=14cm,height=5.3cm]{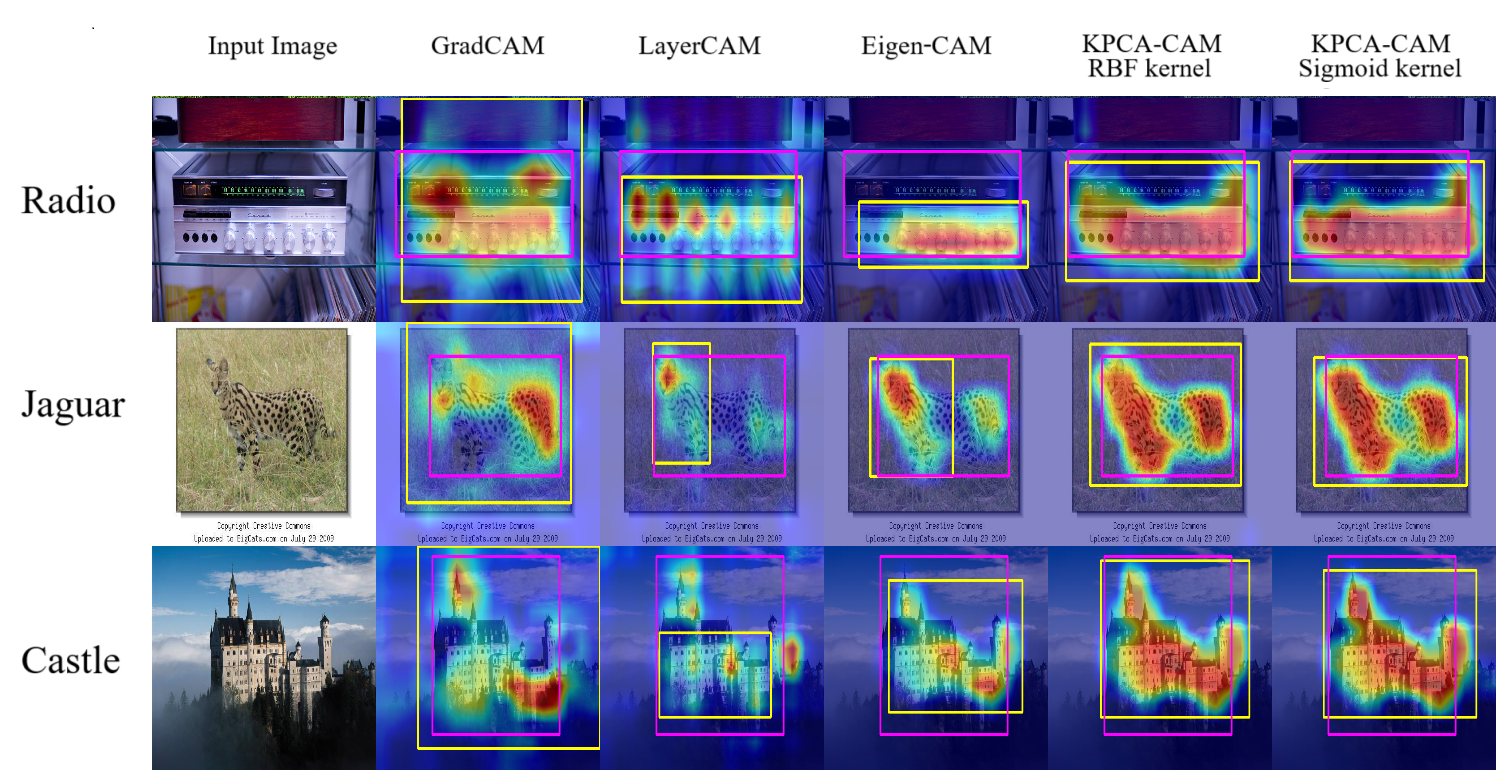}}
\caption{Localization output of different CAMs. The pink box is the ground-truth box from ILSVRC  \cite{ilsvrc} and the yellow box is the predicted bounding box.}
\label{figure_WSOL}
\end{figure*}

This experiment involves creating a bounding box of the image using only the class activation map and checking the IoU between this predicted bounding box and the actual coordinates. IoU \cite{IoU} measures the overlap between two bounding boxes or regions. It is calculated by dividing the area of intersection between the predicted bounding box and the ground truth bounding box by the area of their union. The resulting value lies between 0 and 1, where 0 indicates no overlap and 1 indicates perfect overlap. The mathematical formulation for IoU is

\begin{equation}
IoU = \frac{{\text{{Area of Intersection}}}}{{\text{{Area of Union}}}}. \label{iou}
\end{equation}

In \eqref{iou}, Area of Intersection is the area shared by the predicted bounding box and the ground truth bounding box while Area of Union is the total area covered by both boxes.

\begin{table}[h]
\caption{Localization accuracy (\(\uparrow\)) of CAM techniques }
\begin{tabular}{|l|l|c|c|c|}
\hline
\textbf{Algorithm}         & \textbf{\begin{tabular}[c]{@{}l@{}}Prediction \\ Level\end{tabular}} & \multicolumn{1}{l|}{\textbf{VGG-16}} & \multicolumn{1}{l|}{\textbf{ResNet-50}} & \multicolumn{1}{l|}{\textbf{DenseNet-161}} \\ \hline
\multirow{2}{*}{Grad-CAM}    & loc1                      & 47.94                                  & 69.54                                     & 61.49                                        \\  
                             & loc5                      & 47.07                                  & 68.38                                     & 60.38                                        \\ \hline
\multirow{2}{*}{Grad-CAM++}  & loc1                      & 63.96                                  & 70.71                                     & 60.96                                        \\  
                             & loc5                      & 63.27                                  & 69.77                                     & 60.17                                        \\ \hline
\multirow{2}{*}{Layer-CAM}   & loc1                      & 34.38                                  & 69.21                                     & 66.21                                        \\ 
                             & loc5                      & 34.16                                  & 68.29                                     & 65.11                                        \\ \hline
\multirow{2}{*}{Eigen-CAM}   & loc1                      & 28.32                                  & 55.08                                     & 39.80                                        \\ 
                             & loc5                      & 27.27                                  & 53.46                                     & 39.05                                        \\ \hlineB{2}
\multirow{2}{*}{\textbf{KPCA-CAM(S)}}  & loc1                      & \textbf{72.46}                                  & \textbf{72.35}                                     & \textbf{69.27}                                        \\ 
                             & loc5                      & \textbf{71.01}                                  & \textbf{71.18}                                     & \textbf{67.69}                                        \\ \hline
\multirow{2}{*}{KPCA-CAM(R)}  & loc1                      & 63.48                                  & 63.95                                     & 58.13                                        \\ 
                             & loc5                      & 61.84                                  & 62.58                                     & 57.07                                        \\ \hline
\end{tabular}
\label{results_iou}
\end{table}
    
To create the bounding box for the object using class activation maps, we binarize the map with a threshold of 15\% of the highest value (i.e. 255) and find the largest contour. Using this contour, we generate a compact bounding box. We compute the IoU between this bounding box and the true label. The predicted bounding box is labeled as correct if there is an overlap of more than 50\% (i.e. IoU $\geq$ 0.5).

We evaluate IoU for loc1 and loc5 predictions. loc1 denotes that the model has correctly predicted the category of the image, while loc5 denotes that the target class falls within the top 5 classes predicted by the model.

In Table \ref{results_iou}, we see the results for the various CAM techniques on the weakly supervised localization task. KPCA-CAM with sigmoid kernel outperforms all other CAM techniques. This can be further deduced in Fig. \ref{figure_WSOL}. Eigen-CAM always makes conservative heatmaps, where it accurately locates the object, but fails to capture all the features of the objects. In comparison, Grad-CAM does a better job by generating a broader heatmap, but this also contains noise which leads to a lower IoU. Both KPCA-CAMs circumvent these flaws by emphasizing the significant parts of the object while managing to avoid capturing background objects in the heatmap. KPCA-CAM with the RBF kernel underperforms because it can overfit, picking up noise and missing key features, which lowers the IoU.

\subsubsection{Image Occlusion}
\label{sec:occlusion}

In order to check the significance of the region highlighted by the class-activation map, we computed the Most Relevant First (MoRF) metric \cite{road} for each CAM technique. This metric aims to measure the contribution of each feature to the final decision made by a neural network. Salient pixels selected by the CAM are masked and the new masked image is passed to the model for inference. In the ideal scenario, we should see a drop in confidence for the originally predicted class. Pixels are replaced with a weighted average of their neighboring pixels. This process accounts for potential dependencies among neighboring pixels requiring removal, resulting in a system of linear equations that necessitates solving to determine the updated pixel values. The result can be seen in Fig. \ref{figure_road_bear}.

\begin{figure}[h]
\centerline{\includegraphics[width=8cm,height=4cm]{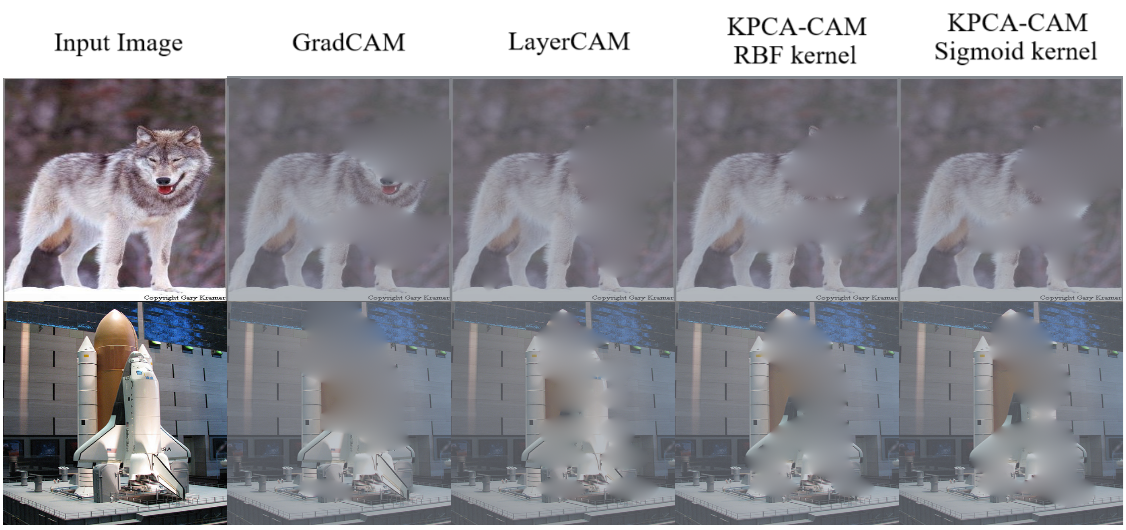}}
\caption{Blurred images generated by masking the salient features of the target object determined by different CAM algorithms}
\label{road_cam}
\end{figure}

For all images correctly classified by the model, 25\% of the significant region determined by the CAM algorithm is blurred. These blurred images, as seen in Fig. \ref{road_cam}, are passed to the model to get a new confidence score for the object targets. The ROAD Most Relevant metric takes the difference between the new confidence scores with the original. A lower ROAD value signifies that the CAM produced by the respective algorithm highlighted salient features that were necessary to identify the target class. 

It can be seen in Table \ref{results_road} that the KPCA-CAMs have comparable ROAD values with all other CAM techniques. Other CAMs have lower ROAD values because their class activation maps are generated by gradients, therefore removing those regions directly lowers the activations of fully connected layers which decreases the confidence score of the prediction. KPCA-CAM achieves a similar ROAD score without any feedback from the fully-connected layers.

\begin{table}[h]
\centering
\caption{ROAD value (\(\downarrow\)) of CAM techniques }
\begin{tabular}{|l|l|l|l|}
\hline
\textbf{Algorithm} & \textbf{VGG-16} & \textbf{ResNet-50} & \textbf{DenseNet-161}          \\ \hline
Grad-CAM           & -55.37           & \textbf{-46.78}              & \textbf{-37.10}                \\ \hline
Grad-CAM++         & -55.55           & -46.69              & -36.91                \\ \hline
Layer-CAM          & \textbf{-57.15}           & -46.24              & -36.89                \\ \hline
Eigen-CAM          & -54.84           & -44.18              & -33.67                \\ \hlineB{2}
KPCA-CAM(S)        & -55.05           & -44.21              & -32.30                \\ \hline
KPCA-CAM(R)        & -54.41           & -37.55              & -26.88                \\ \hline
\end{tabular}
\label{results_road}
\end{table}

\section{Conclusion}
\label{sec:conclusion}

In this paper, we present a novel approach to Class Activation Mapping, KPCA-CAM, which uses advanced kernel techniques to enhance the interpretability of convolutional deep neural networks. Through rigorous experiments, we demonstrate the effectiveness of KPCA-CAM in providing clearer visual explanations of model predictions. KPCA-CAM is intuitive and easy to use, requiring only the learned representations from the last convolution layer, making it independent of fully connected layers. These insights aid in understanding neural network decision-making and improving computer vision model performance. Future work in this field can be to explore the use of CAM techniques for explainability of vision models in tasks such as caption generation and visual question answering.

\section*{Acknowledgment}

We are grateful to Professor Zhou Wang in the Electrical and Computer Engineering department at the University of Waterloo for their guidance and the resources they provided to conduct our research.

\bibliographystyle{IEEEbib}
\bibliography{publication}

\end{document}